\documentclass[11pt,a4paper]{article}
\usepackage{times,latexsym}
\usepackage{url}
\usepackage[T1]{fontenc}

\usepackage[acceptedWithA]{tacl2018v2}

\usepackage{amsmath}
\usepackage{amsfonts}
\usepackage{amssymb}
\usepackage{wrapfig}
\usepackage{subcaption}
\usepackage{multirow}
 \usepackage{mathtools} 

\usepackage{verbatim}

\usepackage{anyfontsize}

\usepackage{color}
\usepackage{tikz}
\usetikzlibrary{arrows,shapes,snakes,automata,backgrounds,fit,petri}
\usepackage{adjustbox}

\newcommand{\RN}[1]{%
	\textup{\lowercase\expandafter{\it \romannumeral#1}}%
}

\usepackage{csquotes}
\usepackage{booktabs}

\usepackage{tcolorbox}
\usepackage{colortbl}

\usepackage{xcolor}
\definecolor{mygreen}{HTML}{3cb44b}
\colorlet{myyellow}{green!10!orange!90!}
\makeatletter
\newcommand{\distas}[1]{\mathbin{\overset{#1}{\kern\z@\sim}}}%

\usepackage{enumitem}

\usepackage[lined,boxed,commentsnumbered,ruled,linesnumbered]{algorithm2e}

\newcommand{\ie}[0]{\emph{i.e., }}

\newcommand{\eg}[0]{\emph{e.g., }}

\newcommand{\beq}{\vspace{0mm}\begin{equation}}
\newcommand{\eeq}{\vspace{0mm}\end{equation}}
\newcommand{\beqs}{\vspace{0mm}\begin{eqnarray}}
\newcommand{\eeqs}{\vspace{0mm}\end{eqnarray}}
\newcommand{\barr}{\begin{array}}
\newcommand{\earr}{\end{array}}

\newcommand{\bv}[0]{{\boldsymbol{b}}}
\newcommand{\cv}[0]{{\boldsymbol{c}}}

\newcommand{\rv}{\boldsymbol{r}}
\newcommand{\sv}[0]{{\boldsymbol{s}}}

\newcommand{\xv}{\boldsymbol{x}}

\newcommand{\thetav}{\boldsymbol{\theta}}

\newcommand{\Lcal}{\mathcal{L}}

\newcommand{\Dcal}{\mathcal{D}}

\ifx\assumption\undefined
\newtheorem{assumption}{Assumption}
\fi

\ifx\definition\undefined
\newtheorem{definition}{Definition}
\fi

\ifx\remark\undefined
\newtheorem{remark}{Remark}
\fi

\usepackage{color, colortbl}
\definecolor{Gray}{gray}{0.93}

\usepackage{xspace,mfirstuc,tabulary}

\newif\iftaclinstructions
\taclinstructionsfalse 
\iftaclinstructions
\renewcommand{\confidential}{}
\renewcommand{\anonsubtext}{(No author info supplied here, for consistency with
TACL-submission anonymization requirements)}
\newcommand{\instr}
\fi

\iftaclpubformat 

\else

\fi

\title{\model{}: Building Task Bots at Scale with \\ Transfer Learning and Machine Teaching}

\author{Baolin Peng, Chunyuan Li, Jinchao Li \\
{\bf Shahin Shayandeh, Lars Liden, Jianfeng Gao} \\
  Microsoft Research, Redmond \\
  \texttt{\{bapeng,chunyl,jincli,shahins,lars.liden,jfgao\}@microsoft.com}
  }

\date{}

\newcommand{\model}{\textsc{Soloist}}
\newcommand{\modelp}[1]{\model{$_{\mathtt{#1}}$}}
\newcommand{\longname}{\textsc{Ta\textbf{S}k-\textbf{O}riented Dia\textbf{LO}g w\textbf{I}th a \textbf{S}ingle pre-\textbf{T}rained model}}

\begin{document}
\maketitle
\begin{abstract}
We present a new method \model{}\footnote{\longname{}. In this paper, \model{} refers to both the proposed bot building method and the dialog model or system developed using the method.} that uses transfer learning and machine teaching to build task bots at scale. 
We parameterize classical modular task-oriented dialog systems using a Transformer-based auto-regressive language model, which subsumes different dialog modules into a single neural model.
We pre-train, on heterogeneous dialog corpora, a task-grounded response generation model, which can generate dialog responses grounded in user goals and real-world knowledge for task completion. 
The pre-trained model can be efficiently adapted to accomplish new tasks with a handful of task-specific dialogs via machine teaching,
where training
samples are generated by human teachers interacting with the system.
Experiments show that 
$(\RN{1})$ \model{} creates new state-of-the-art on well-studied task-oriented dialog benchmarks, including CamRest676 and MultiWOZ;
$(\RN{2})$ in the few-shot fine-tuning settings, \model{} significantly
outperforms existing methods, and 
$(\RN{3})$ the use of machine teaching substantially reduces the labeling cost of fine-tuning. The pre-trained models and codes are available at \url{https://aka.ms/soloist}.
\end{abstract}

\begin{figure*}[t!]
\centering
\begin{tabular}{c}
\includegraphics[width=1.85\columnwidth]{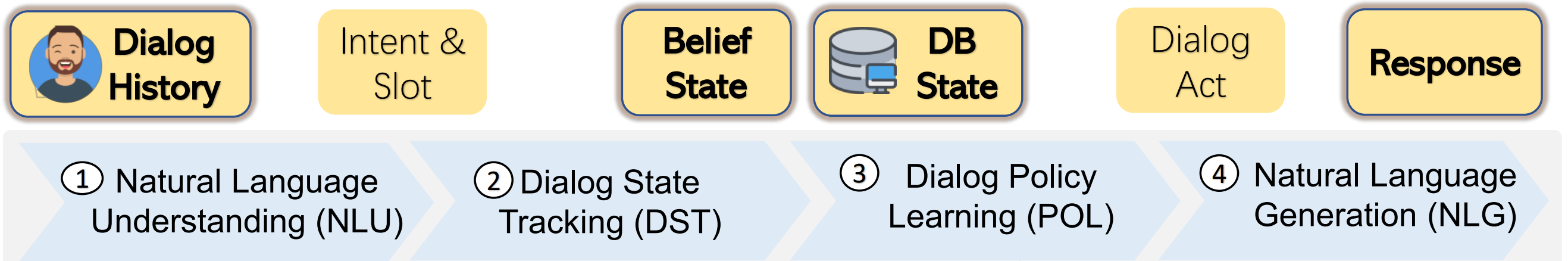}\\
(a) A typical task-oriented dialog system pipeline. \vspace{2mm}\\
\includegraphics[width=1.85\columnwidth]{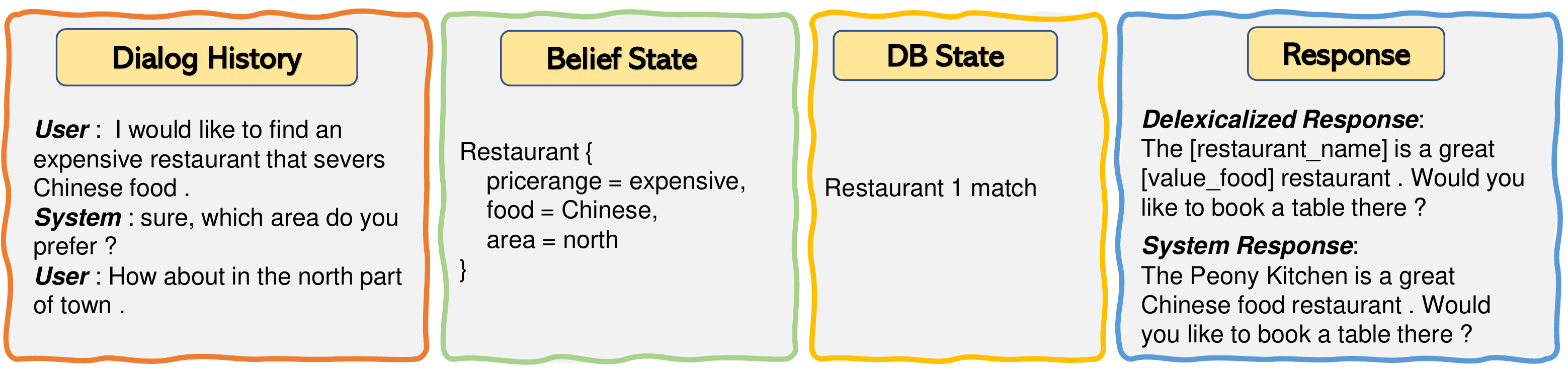}\\
(b) Example snippets for the items compounding the input of \model{} model. \vspace{2mm} \\
\includegraphics[width=1.85\columnwidth]{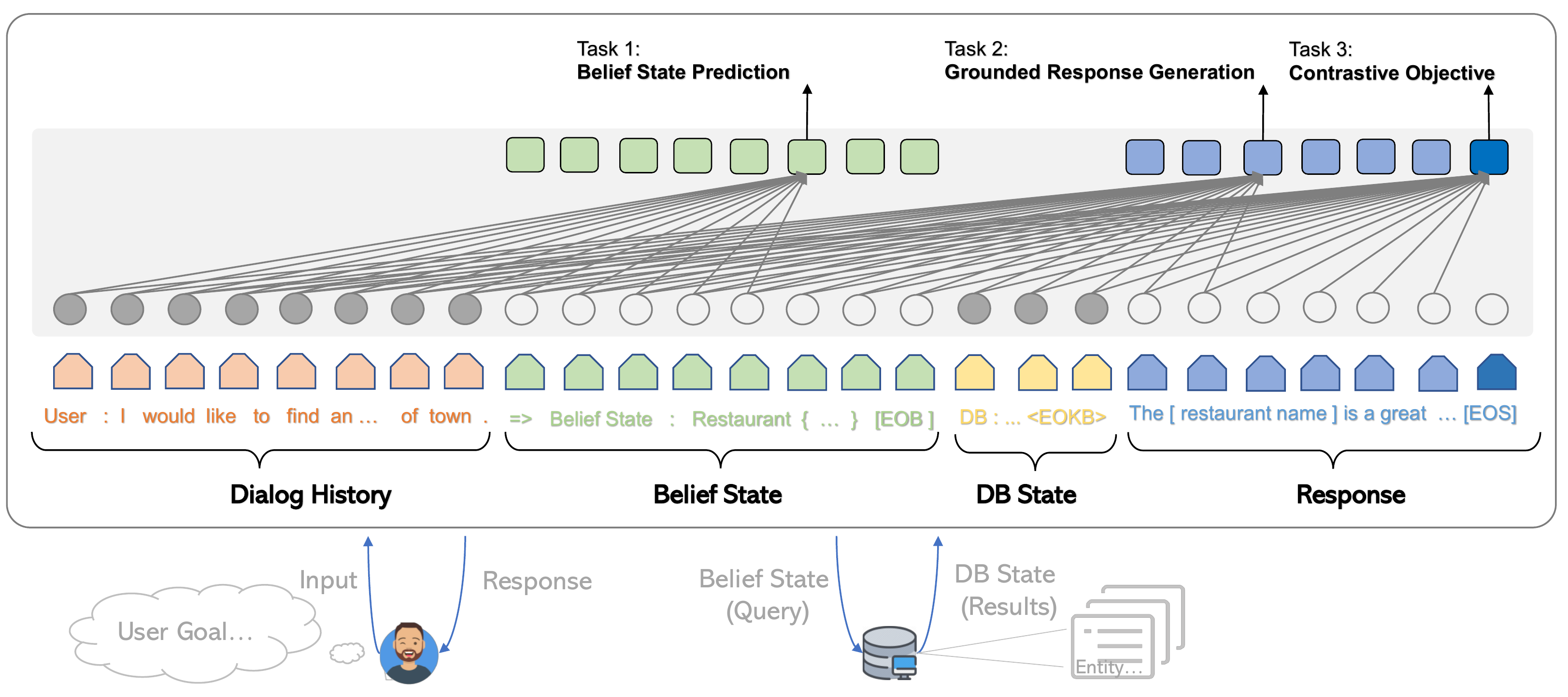}\\
(c) The proposed \model{} model architecture and training objectives.
\end{tabular}
\caption{Illustration of a traditional modular task-oriented dialog system, an example for the model input, and the proposed model. The  \model{} solution utilizes a single neural auto-regressive model in (c) to paramterize the sequential dialog pipeline in (a), with input sequence represented in (b). Different from GPT-2, the \model{} model learns to ground response generation in user goals and database/knowledge.}

\label{fig:task_dialog_scheme}
\end{figure*}

\section{Introduction}
The increasing use of personal assistants and messaging applications has spurred interest in building task-oriented dialog systems (or task bots) that can communicate with users through natural language to accomplish a wide range of tasks, such as restaurant booking, weather query, flight booking, IT helpdesk \cite[\eg][]{DBLP:journals/coling/ZhouGLS20,adiwardana2020towards,roller2020recipes,DBLP:journals/corr/abs-2009-03457,DBLP:journals/corr/abs-2012-14666}. 
The wide variety of tasks and domains has created the need for a flexible task-oriented dialog development platform that can support many different use cases while remaining straightforward for developers to use and maintain.

A typical task-oriented dialog system uses a modular pipeline, which has four modules and executes sequentially~\cite{DBLP:journals/pieee/YoungGTW13,gao2019neural}, as shown in Figure~\ref{fig:task_dialog_scheme} (a).
A natural language understanding ({NLU}) module identifies user intents and extracts associated information such as slots and their values from user’s input. A dialog state tracker ({DST}) infers the belief state (or user goal) from dialog history.
The belief state is often used to query a task-specific database (DB) to obtain the DB state, such as the number of entities that match the user goal.
The dialog state and DB state are then passed to a dialog policy ({POL}) to select the next system action.
A natural language generation ({NLG}) module converts the action to a natural language response.

Most popular commercial tools for dialog development employ the modular systems,
including 
Google's Dialog Flow\footnote{{https://dialogflow.com/}},
Microsoft's Power Virtual Agents (PVA)\footnote{{https://powervirtualagents.microsoft.com/}}, 
Facebook's Wit.ai\footnote{{https://wit.ai/}}, 
Amazon's Lex\footnote{{https://aws.amazon.com/lex/}}, and 
IBM's Watson Assistant\footnote{{https://www.ibm.com/watson/}}.
They are designed mainly to help develop systems \emph{manually}, \ie writing code, crafting rules and templates.  
Unfortunately, even with these tools, building dialog systems remains a label-intensive, time-consuming task, requiring rich domain knowledge, reasonable coding skill, and expert experience. 
The cost of building dialog systems at scale (\ie tens of thousands of bots for different tasks)
can be prohibitively expensive.

With the recent advances in neural approaches to conversational AI~\cite{gao2019neural}, researchers have been developing data-driven methods and neural models for either individual dialog modules or end-to-end systems.
For example, recent attempts such as RASA \cite{rasa}, ConvLab \cite{convlab, convlab-2}, and Conversation Learner~\cite{shukla2020conversation} are made to allow the use of data-driven approaches based on machine learning and machine teaching to develop dialog modules. 
End-to-end trainable dialog systems have also been studied~\cite[\eg][]{wen2016network,zhao2016towards,li2017end,williams2017hybrid,lei2018sequicity, gao2019neural, zhang2019task}. Although these methods have achieved promising results, they require large amounts of task-specific labeled data for training, which are rarely available for new tasks in real-world applications.

In this paper we propose a novel method of building task bots at scale, \model{}, which significantly eases the workflow of training and deploying dialog systems for new tasks, compared to existing tools and methods.
Our approach is inspired by the recent success of applying transfer learning to natural language processing (NLP) tasks:
Big language models pre-trained on large amounts of raw text (\eg BERT~\cite{devlin2019bert}, RoBERTa~\cite{liu2019roberta} and UniLM~\cite{dong2019unified}) can be effectively fine-tuned for a wide range of NLP tasks with few in-domain labels. 
Recently, these pre-trained language models have also been employed to develop dialog modules such as NLU and DST ~\cite{DBLP:conf/emnlp/HendersonCMSWV20,DBLP:conf/acl/CoopeFGVH20,Wu2020ToDBERTPN}. 
The proposed \model{} uses a similar pre-training-and-fine-tuning framework for building end-to-end dialog systems.
We parameterize a task bot using a Transformer-based auto-regressive language model, which subsumes different dialog modules (\ie NLU, DST, POL and NLG) into a single neural model.
Task bot building proceeds in two stages: 
$(\RN{1})$ In the pre-training stage, initialized using GPT-2~\cite{gpt2}, we train a Transformer-based, task-grounded, response generation model using large heterogeneous dialog corpora. The model learns the primary task completion skills such as DST and POL, and can generate dialog responses \emph{grounded} in user goals and real-world knowledge for task completion. 
$(\RN{2})$ In the fine-tuning stage, we adapt the pre-trained \model{} model to complete a specific (new) task using a handful of task-specific dialogs via machine teaching, where training samples are generated by human teachers interacting with the system~\cite{zhu2015machine,shukla2020conversation}.

We show through a comprehensive empirical study that \model{} is an effective method of building task bots at scale by successfully transferring two capabilities from the pre-trained model to a new task bot: 
$(\RN{1})$ the capability of NLU and NLG
learned on raw text, and 
$(\RN{2})$ the capability of grounding system responses in user goals and real-world knowledge for task completion, learned on the out-domain dialog corpora.

\model{} achieves state-of-the-art performance on two well-studied task-oriented dialog benchmarks, lifting the combined score by 10 points in automatic evaluation, and the success rate by 20 points in human evaluation.
In the few-shot fine-tuning settings, \model{} adapts to the new domain much more effectively than competing methods, achieving a reasonable success rate using less than 50 dialogs.
The promising results demonstrate the potential of the new method for developing task bots at scale.
Instead of collecting, labeling data, and building one bot per task, we can pre-train a task-grounded response generation model, and adapt it to new tasks via transfer learning and machine teaching.

\section{\model{}}

\subsection{An Auto-Regressive Model for Dialog}
The modular dialog system in Figure~\ref{fig:task_dialog_scheme} constitutes a data processing pipeline that produces a sequence, through concatenating the input-output pair of each module along the generation process. Each consecutive pair in this sequence plays the role of annotated data for the corresponding module. Ideally, when the entire sequence is available, the data generation process of a dialog system (NLU, DST, POL, NLG) can be formulated as a {\em single} auto-regressive model.

GPT-2~\cite{gpt2} is a state-of-the-art (SoTA) auto-regressive language model
trained on large amounts of open Web text data.
Although after being fine-tuned using conversational data, GPT-2 can respond to users with realistic and coherent continuations about any topic of their choosing~\cite{zhang2019dialogpt}, the generated responses are not useful for completing any specific task due to the lack of grounding.
\model{} inherits GPT-2's capability of producing human-like responses. Nevertheless, unlike GPT-2, \model{} is pre-trained to generate responses grounded in user goals and real-world knowledge for task completion. While GPT-2 is a language model for text prediction, \model{} is a stateful decision-making model for task completion, with the capabilities of tracking dialog states, selecting best system actions, and so on. 
Thus, \model{} is pre-trained using task-oriented dialog sessions annotated with grounding information, \ie user goals, dialog belief states, database (DB) states, and system responses. 
Specifically, each dialog turn in our training data is represented as:
\begin{align}
\vspace{-2mm}
\label{eq_lm_input}
\xv = ( \sv, \bv, \cv, \rv ), 
\vspace{-2mm}
\end{align}
where $\sv$ is the dialog history up to the current dialog turn, $\bv$ is the dialog belief state acquired from human annotation, $\cv$ is the DB state automatically retrieved from a database using $\bv$, and $\rv$ is the delexicalized dialog response, from which the system response in natural language can be generated using some automatic post-processing. 
Each item in $\xv$ is by itself a sequence of tokens, as illustrated by the examples in Figure~\ref{fig:task_dialog_scheme} (b).
Thus, it is natural to treat the concatenation of them as a long sequence for model training as shown in Figure~\ref{fig:task_dialog_scheme} (c). 
We pre-train the \model{} model using publicly available heterogeneous dialog corpora with labels of belief states and DB states.
The pre-trained model can be fine-tuned to any new task to generate responses grounded in task-specific user goals and a database.

\subsection{Task-Grounded Pre-Training}
\label{sec:method}

Given training data of $N$ samples $\Dcal=\{ \xv_n\}_{n=1}^{N}$, 
our goal is to build a neural model parameterized by $\thetav$ to characterize the sequence generation probability $p_{\thetav}(\xv)$. We use a multi-task objective for learning $\thetav$, where each task is a self-supervised learning task. 

To leverage the sequential structure of a task-oriented dialog system, the joint probability $p(\xv)$ can be factorized in the auto-regressive manner as:
\begin{align}
\label{eq_factorization}
p(\xv) & = p(\rv, \cv, \bv, \sv) \\
& = \underbrace{p(\rv | \cv, \bv, \sv)}_{\text{\hspace{-14mm}Grounded Response Generation}}
\hspace{-8mm} 
\underbrace{p(\bv | \sv)}_{\text{\hspace{10mm}  Belief Prediction}}
\hspace{-8mm} 
p(\sv),
\label{eq_factorization2}
\end{align}
where the factorization from \eqref{eq_factorization} to \eqref{eq_factorization2} is based on the fact that $p(\cv |\bv, \sv) = p(\cv | \bv) = 1 $, because the DB state $\cv$ is obtained using a deterministic database-lookup process given a belief state $\bv$ (\eg via an API call).    
Note that \eqref{eq_factorization2} decomposes the joint distribution modeling problem into two sub-problems: belief state prediction $p(\bv | \sv)$ and grounded response generation $p(\rv | \cv, \bv, \sv)$. Since $\bv $ and $\rv $ are sequences, we can further factorize them in the left-to-right auto-regressive manner, respectively.  

\paragraph{Task 1: Belief Prediction.}
For a belief state sequence of length $T_b$, we define the objective of predicting the belief state as: 
\begin{align}
\vspace{-2mm}
\label{eq_lm_belief}
\Lcal_{\text{B}} = 
\log p(\bv | \sv ) = \sum_{t=1}^{T_b} \log p_{\thetav}(b_t| b_{<t}, \sv),
\vspace{-2mm}
\end{align}
where $b_{<t}$ indicates all tokens before $t$.

\paragraph{Task 2: Grounded Response Generation.}
A delexicalized response of length $T_r$, $\rv =[r_1, \cdots, r_{T_r}]$, is generated by our model token-by-token from left to right, grounded in dialog history $\cv$, belief state $\bv$ and DB state $\sv$.
The corresponding training objective is defined as
\begin{align}
\vspace{-2mm}
\label{eq_lm_response}
\Lcal_{\text{R}} & = 
\log p(\rv | \cv, \bv, \sv ) \\
&  = \sum_{t=1}^{T_r} \log p_{\thetav}(r_t| r_{<t}, \cv, \bv, \sv).  
\nonumber
\vspace{-2mm}
\end{align}
%

\paragraph{Task 3: Contrastive Objective.} 
A contrastive objective is employed to promote the matched items (positive samples $\xv$) while driving down the mismatched items (negative samples $\xv^{\prime}$).  
The negative samples are generated from sequence $\xv$ by replacing some items in $\xv$ with probability 50\% with different items randomly sampled from the dataset $\Dcal$.  
Since the the special token $\texttt{[EOS]}$ attends all tokens in the sequence, the output feature on $\texttt{[EOS]}$ is the fused representation of all items. We apply a binary classifier on top of the feature to predict whether the items in the sequence are matched ($y=1$) or mismatched ($y=0$). The contrastive object is cross-entropy defined as:
\begin{align}
\vspace{-2mm}
\label{eq_ranking}
\hspace{-3mm}
\Lcal_{\text{C}} \!=\! y \log(p_{\thetav}(\xv)) + (1\!-\!y) \log (1-p_{\thetav}(\xv^{\prime})).
\vspace{-2mm}
\end{align}
We generate three types of negative samples $\xv^{\prime}$, each of which is chosen with probability 1/3:
$(\RN{1})$
{\em negative belief}, where only the belief state item is replaced
$(\RN{2})$
{\em negative response}, where only the response item is replaced
$(\RN{3})$
{\em negative belief + response}, where both the belief state and response items are replaced.

\paragraph{Full Pre-Training Objective.} 
$\thetav$ is learned via maximizing the log-likelihood over the training dataset $\Dcal$, using a joint objective that combines~\eqref{eq_lm_belief}, \eqref{eq_lm_response} and \eqref{eq_ranking}:
\begin{equation}\label{eq_multi_task}
    \Lcal_{\thetav}(\Dcal) =  \sum_{n=1}^{|\Dcal|}  
    (\Lcal_{\text{B}}(\xv_n) + \Lcal_{\text{R}}(\xv_n) + \Lcal_{\text{C}}(\xv_n)).
\end{equation}
Figure~\ref{fig:task_dialog_scheme} (c) illustrates the model architecture and learning objectives. The model is auto-regressive in a left-to-right manner, with each of the three training tasks labeled on its corresponding output (\ie sub-sequence separated by a special token). 

\paragraph{Implementation Details.} 
Each dialog turn in training data is processed to form a sequence of tokens consisting of four items $( \sv, \bv, \cv, \rv )$. For example, the dialog turn of Figure~\ref{fig:task_dialog_scheme} (b) is represented as follows, where different items are rendered in different colors.
\begin{tcolorbox}[boxsep=1pt,colframe=white!90!black,colback=black!2!white] \small
    \textcolor{red!60}{User : I would like to find an expensive restaurant that severs Chinese food. System : sure, which area do you prefer ? User :  How about in the north part of town.} 
    \textcolor{mygreen}{$\mathtt{=>}$ Belief State : Restaurant \{ pricerange = expensive, food = Chinese, area = north \} $\mathtt{<EOB>}$}
    \textcolor{myyellow}{DB : Restaurant 1 match $\mathtt{<EOKB>}$}
    \textcolor{blue}{The [restaurant\_name] is a great [value\_food] restaurant. Would you like to book a table there ? $\mathtt{<EOS>}$}
\end{tcolorbox} 

\begin{table}[t!]
    \centering
    \footnotesize
    
    \setlength{\tabcolsep}{1.0mm}{
    \begin{tabular}{lcccc}
    \toprule
Name & \#Dialog & \#Utterance & Avg. Turn & \#Domain \\
\midrule
\multicolumn{5}{l}{{\hspace{-1mm} \em task-grounded pre-training: }}  \\
Schema & 22,825 & 463,284 & 20.3 & 17 \\
Taskmaster & 13,215 & 303,066 & 22.9 & 6 \\
\midrule
\multicolumn{5}{l}{{\hspace{-1mm} \em fine-tuning: }}  \\
MultiWOZ2.0 & 10,420  & 71,410 & 6.9 & 7 \\
CamRest676 & 676 & 2,744 & 4.1& 1 \\
Banking77 & - & 25,716 & - & 21 \\
Restaurant-8k & - & 8,198 & - & 1 \\

    \bottomrule
    \end{tabular}
    }
    \caption{Dialog corpora. The datasets in the upper block are used for task-grounded pre-training, and the datasets in the lower block are for fine-tuning.}
    \label{table:datasetstatistics}
    \vspace{-5mm}
\end{table}
\begin{figure*}[t!]~\hspace{3mm}
\includegraphics[width=2\columnwidth]{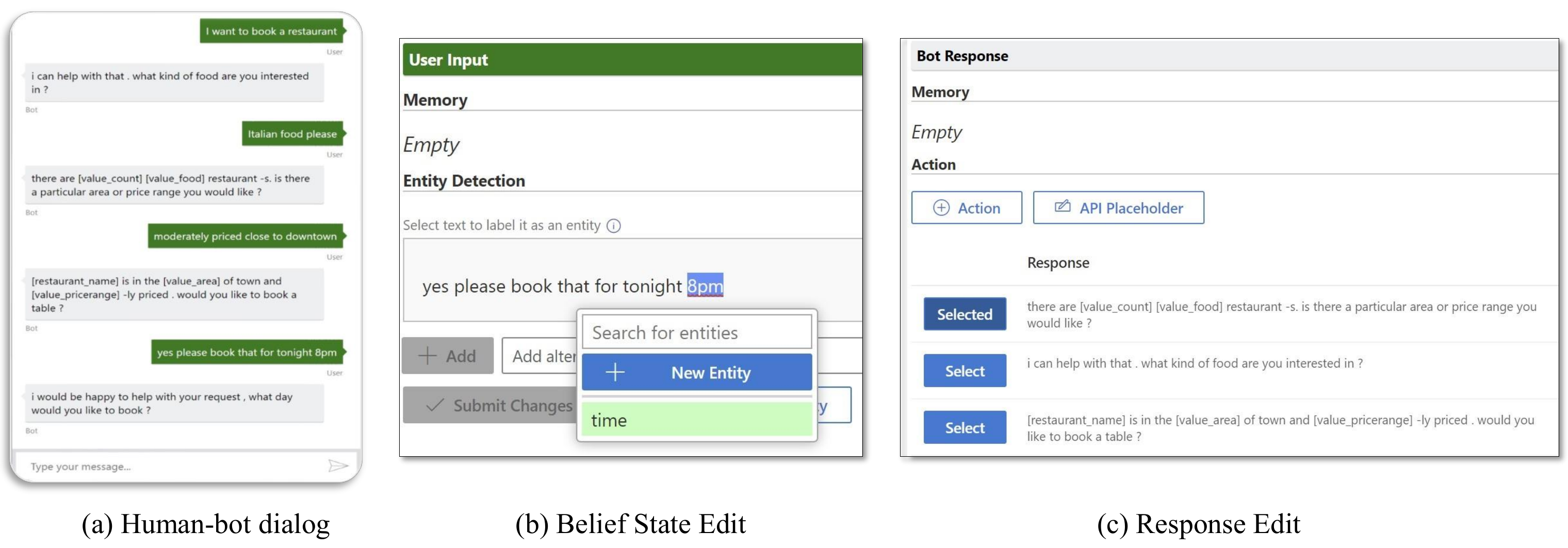}
\caption{Illustration of the machine teaching process using conversion learner. The human-bot conversion log in (a) can be edited via correcting its belief state in (b), and selecting/inserting a more appropriate response in (c).}
\label{fig:example_mt}
\end{figure*}
 
This sequence, tokenized using byte pair encodings~\citep{bpe}, can be readily used for multi-task training,
as shown in Figure~\ref{fig:task_dialog_scheme}(c). 
The implementation of \model{} is based on Huggingface Pytorch Transformer \cite{Wolf2019HuggingFacesTS}. 
The task-grounded pre-training of \model{} uses the public 117M-parameter GPT-2 as initialization. 
Adam~\cite{kingma2014adam} with weight decay is used for pre-training. 
Table \ref{table:datasetstatistics} shows the dialog corpora \cite{kim2019eighth,rastogi2019towards,byrne2019taskmaster} used for task-grounded pre-training.  
To ensure there is no overlap between pre-training and fine-tuning datasets, we exclude the data akin to MultiWOZ~\cite{budzianowski2018multiwoz}, CamRest676 \cite{wen2016network}, Banking77 \cite{casanueva2020efficient}, Restaurant-8k \cite{DBLP:conf/acl/CoopeFGVH20}. 

\subsection{Fine-Tuning and Machine Teaching}

When deploying \model{} to a new task, we collect task-specific $\xv$ in the same format as that used for pre-training as~\eqref{eq_lm_input}.
When $\xv$ is available, the conventional fine-tuning procedure is utilized: we use the same multi-task objective of \eqref{eq_multi_task} to update $\thetav$ to adapt the model to complete the new task using labeled task-specific dialogs. 

In real applications, annotated task-specific data is often unavailable, or noisy/incomplete beforehand. One may deploy the dialog system, and acquire high-quality task-specific labels (\eg belief state and system response) for each dialog turn using machine teaching.
Machine teaching is an active learning paradigm that focuses on leveraging the knowledge and expertise of domain experts as ``teachers''. This paradigm puts a
strong emphasis on tools and techniques that enable teachers - particularly non-data scientists and
non-machine-learning experts - to visualize data,
find potential problems, and provide corrections or
additional training inputs in order to improve the
system’s performance~\citep{machine-teaching-2017,zhu2015machine,williams2017demonstration,shukla2020conversation}.

We proceed fine-tuning using Conversation Learner~\cite{shukla2020conversation}, a machine teaching tool, in the following steps: 
$(\RN{1})$
Dialog authors deploy the pre-trained \model{} model for a specific task.
$(\RN{2})$
Users (or human subjects recruited for system fine-tuning) interact with the system and generate human-bot dialog logs.
$(\RN{3})$
Dialog authors revise a dozen of training samples by selecting
representative failed dialogs from the logs, correcting their belief and/or responses so that the system can complete these dialogs successfully, as illustrated in Figure~\ref{fig:example_mt}.
The corrected task-specific dialog turns are used to fine-tune the model. 

\paragraph{Implementation Details.} 
To adapt a pre-trained \model{} to a new task in our experiments, we always fine-tune \model{} using a small amount of pre-collected task-specific dialogs, and then continue to fine-tune it via machine teaching, as detailed in Section~\ref{sec:fewshot}. 
Training examples are truncated to ensure a maximal length of 512. The pre-trained models are fine-tuned with a mini-batch of 6 on 8 Nvidia V100 until no progress is observed on validation data or up to 10 epochs. Nucleus sampling~\cite{holtzman2019curious} is used for decoding, where the sampling top-p ranges from 0.2 to 0.5 for all our models. The best setup of hyper-parameters is selected through grid-search on the validation set. For the machine teaching experiment,pre-trained models are fine-tuned with SGD on a single Nvidia V100.

\section{Experiments}

\begin{table*}[t!]
    \centering
    \scriptsize
    
    \setlength{\tabcolsep}{1.0mm}{
    \begin{tabular}{lcccccc}
    \toprule
    
    \multirow{2}{*}{Model} &
\multicolumn{2}{c}{Annotations} &
\multicolumn{4}{c}{Evaluation Metrics} \\
\cmidrule(l){2-3} \cmidrule(l){4-7}
 & Belief State & Policy & $\mathtt{Inform} \uparrow$ & $\mathtt{Success} \uparrow$ & $\mathtt{BLEU} \uparrow$ & $\mathtt{Combined} \uparrow$  \\
\midrule
Sequicity \citep{lei2018sequicity}  & \checkmark & \checkmark & 92.30 & 85.30 & 21.40 & 110.20 \\
Sequicity (w/o RL) & \checkmark & \checkmark & 94.00 & 83.40 & 23.40 & 112.10 \\
GPT fine-tuning \cite{budzianowski2019hello} &  &  & - & 86.20 & 19.20 & - \\
ARDM$^{1}$ \citep{wu2019alternating}&   & & - & 87.10 & 25.20 & -\\
\rowcolor{Gray}
\model{} & \checkmark & & {\bf94.70} & {\bf87.10} & {\bf25.50} & {\bf116.40} \\
    \bottomrule
    \end{tabular}
    }
    
    {\footnotesize \footnotemark[1]ARDM is not fully E2E, as it requires a rule-based dialog state tracker.}
    
    \caption{End-to-End evaluation on CamRest676. Results of existing methods are from \citet{wu2019alternating}. }
    \label{table:camrest676}
\end{table*}

\begin{table*}[t!]
    \centering
    \scriptsize
    
    \setlength{\tabcolsep}{1.0mm}{
    \begin{tabular}{lcccccc}
    \toprule
    
    \multirow{2}{*}{Model} &
\multicolumn{2}{c}{Annotations} &
\multicolumn{4}{c}{Evaluation Metrics} \\
\cmidrule(l){2-3} \cmidrule(l){4-7}
 & Belief State & Policy & $\mathtt{Inform} \uparrow$ & $\mathtt{Success} \uparrow$ & $\mathtt{BLEU} \uparrow$ & $\mathtt{Combined} \uparrow$  \\
\midrule
Sequicity \citep{lei2018sequicity} & \checkmark & \checkmark & 66.41 & 45.32 & 15.54 & 71.41 \\
HRED-TS \citep{peng2019teacher} & \checkmark & \checkmark & 70.00 & 58.00 & {\bf17.50} & 81.50 \\
Structured Fusion \citep{mehri2019structured} & \checkmark & \checkmark & 73.80 & 58.60 & 16.90 & 83.10 \\
DSTC8 Track 1 Winner \footnotemark[1] \citep{Ham2020e2e} & \checkmark & \checkmark & 73.00 & 62.40 & 16.00 & 83.50 \\
DAMD \citep{zhang2019task} & \checkmark & \checkmark & 76.40 & 60.40 & 16.60 & 85.00 \\
\rowcolor{Gray}
\model{} & \checkmark & & {\bf85.50} & {\bf72.90} & 16.54 & {\bf95.74} \\
    \bottomrule
    \end{tabular}
    }
    
    {\footnotesize \footnotemark[1]The result of DSTC8 Track 1 Winner is produced by adapting their code to our setting.}
    \caption{End-to-end evaluation on MultiWOZ.}
    \label{table:e2emultiwoz}
    
\end{table*}

This section evaluates the proposed \model{} to answer three questions:
\textbf{\texttt{Q1}}: How does \model{} perform on standard benchmarks compared to SoTA methods?
\textbf{\texttt{Q2}}: Does \model{} meet the goal of effectively generalizing to new domains in the few-shot fine-tuning setting?
\textbf{\texttt{Q3}}: how effective machine teaching is for fine-tuning? 
Note that we employ the conventional fine-tuning method {\it without} machine teaching for a fair comparison when studying \textbf{\texttt{Q1}} and \textbf{\texttt{Q2}}. 

\subsection{Experimental Setup}

\paragraph{Dialog Datasets for Fine-Tuning.} We validate the end-to-end dialog system performance of \model{} on two well-studied datasets.
$(\RN{1})$ CamRest676~\cite{wen2016network} is a single-domain task-oriented dialog corpus. It contains 408/136/136 dialogs for training/validation/testing, respectively. Following \citet{lei2018sequicity}, we delexicalize each token that occurs in the ontology with its slot names such as restaurant name, phone number, and postcode. 
$(\RN{2})$ MultiWOZ dataset \cite{budzianowski2018multiwoz} is a multi-domain task-oriented dialog dataset. It contains 8438/1000/1000 for training/validation/testing, respectively. Each dialog session contains 1 to 3 domains, such as Attraction, Hotel, Hospital, Police, Restaurant, Train, and Taxi. MultiWOZ is inherently challenging due to its multi-domain setting and diverse language styles. 

\paragraph{Automatic Evaluation Metrics.} 
Following \citet{budzianowski2018multiwoz}, $\mathtt{Inform}$, $\mathtt{Success}$, and $\mathtt{BLEU}$ scores are reported. 
The first two metrics relate to the dialogue task completion -- whether the system has provided an appropriate entity ($\mathtt{Inform}$) and then answered all the requested attributes ($\mathtt{Success}$).
$\mathtt{BLEU}$ evaluates how natural the generated responses are compared to that generated by human agents.
A combined score ($\mathtt{Combined}$) is also reported using $\mathtt{Combined} = (\mathtt{Inform} + \mathtt{Success}) \times 0.5 + \mathtt{BLEU}$ as an overall quality measure.

\paragraph{Baselines.} We compare \model{} with several strong baselines, which hold SoTA on the CamRest676 or MultiWOZ datasets.
$(\RN{1})$  Multi-Action Data Augmentation (DAMD)~\cite{zhang2019task} is a modular system, where each dialog module is implemented using a neural network, and the whole system is trained in an end-to-end manner. 
$(\RN{2})$ Sequicity \citep{lei2018sequicity} is similar to DAMD except that it does not use multi-action data augmentation.
$(\RN{3})$ GPT fine-tuning~\citep{budzianowski2019hello} is fine-tuned on GPT-2 to generate responses based on the dialog state and history.
$(\RN{4})$ ARDM~\citep{wu2019alternating} utilizes GPT-2 as the pre-trained model to learn to generate role-aware responses given dialog context. The model has to work with a separate dialog state tracker for task completion.
$(\RN{5})$ HDSA~\cite{chen-etal-2019-semantically} is a modular dialog system, which generates responses using a BERT-based dialog policy and graph structure dialog act representations.

\subsection{End-to-End Evaluation}

\paragraph{CamRest676.} 
Table~\ref{table:camrest676} shows the result and lists annotations used by different models. 
\model{} achieves the best scores in all the metrics. 
ARDM performs similarly to \model{} in terms of $\mathtt{Success}$ and $\mathtt{BLEU}$. However, ARDM cannot track dialog states and requires a separately trained state tracker to accomplish tasks. GPT-2 fine-tuned with task-specific data works reasonably well but lags behind \model{} by a large margin. 
Sequicity, which uses a jointly trained model with belief state and policy annotations, underperforms \model{}. 
This result also show that compared to other end-to-end models, \model{} not only achieves better performance but requires lower labeling cost for fine-tuning
due to the use of task-grounded pre-training.

\paragraph{MultiWOZ.}  
The result is shown in Table \ref{table:e2emultiwoz}. 
\model{} achieves the best performance in terms of $\mathtt{Inform}$, $\mathtt{Success}$, and $\mathtt{Combined}$, lifting the previous SoTA by a significant margin (\eg about 10 points improvement in $\mathtt{Combined}$ over DAMD). 
\model{} also outperforms the method of \citet{Ham2020e2e}, where GPT-2 is fine-tuned and applied for end-to-end dialog modeling.
Compared to the classical modular dialog systems such as 
DAMD, \model{} uses a much simpler architecture and requires much lower labeling effort. 
For example, \model{} requires only the belief states, while DAMD requires additional annotations for task definition (\ie defining the intents, slots, and the corresponding value ranges) and dialog acts. 

\begin{table}[t!]
    \centering
    \scriptsize
\scalebox{0.85}{    
    \setlength{\tabcolsep}{1.0mm}{
    \begin{tabular}{lccccc}
    \toprule
    
Domain &\texttt{Attra.} & \texttt{Train}  & \texttt{Hotel}  & \texttt{Rest.} & \texttt{CamRest676} \\
\midrule
\#Train & 50 & 50 & 50 & 50 & 20 \\
\#Valid & 50 & 50 & 50 & 50 & 136 \\
\#Test & 100 & 200 & 200 & 200 & 136\\

    \bottomrule
    \end{tabular}
    }}
    \caption{Data statistics for domains used  in few-shot evaluation. \texttt{Attra.} denotes \texttt{Attraction} domain and \texttt{Rest.} means \texttt{Restaurant}.}
    \label{table:fewshotdataset}
\end{table}

 \begin{table}[t!]
    \centering
    \scriptsize
    
    \setlength{\tabcolsep}{1.0mm}{
    \begin{tabular}{lccccccccccccccc}
    \toprule
    
    \multirow{2}{*}{Model} &
\multicolumn{3}{c}{\texttt{CamRest676}} \\
\cmidrule(l){2-4} 
&$\mathtt{Inform} \uparrow$ & $\mathtt{Success} \uparrow$ & $\mathtt{BLEU} \uparrow$ \\
\midrule
Sequicity \citep{lei2018sequicity} & 60.61 & 66.11 & 11.15 \\
\model{} w/o pre-training & 73.88 & 72.22 & 13.11 \\
\model{} & 85.82 & 84.22 & 19.18 \\
\rowcolor{Gray}
\modelp{L} & {\bf88.05} & {\bf84.79} & {\bf18.88} \\

    \bottomrule
    \end{tabular}
    }
    \caption{End-to-end evaluation on CamRest676 in the few-shot fine-tuning setting.}
    \label{table:fewshotcamrest}
    \vspace{-3mm}
\end{table}

 \begin{table*}[t!]
    \centering
    \tiny
    
    \setlength{\tabcolsep}{1.0mm}{
    \begin{tabular}{lccccccccccccccc}
    \toprule
    
    \multirow{2}{*}{Model} &
\multicolumn{3}{c}{\texttt{Attraction}} &
\multicolumn{3}{c}{\texttt{Train}} &
\multicolumn{3}{c}{\texttt{Hotel}} &
\multicolumn{3}{c}{\texttt{Restaurant}} \\
\cmidrule(l){2-4} \cmidrule(l){5-7} \cmidrule(l){8-10} \cmidrule(l){11-13} \cmidrule(l){14-16} 

&$\mathtt{Inform} \uparrow$ & $\mathtt{Success} \uparrow$ & $\mathtt{BLEU} \uparrow$ & 
$\mathtt{Inform} \uparrow$ & $\mathtt{Success} \uparrow$ & $\mathtt{BLEU} \uparrow$ & 
$\mathtt{Inform} \uparrow$ & $\mathtt{Success} \uparrow$ & $\mathtt{BLEU} \uparrow$ & 
$\mathtt{Inform} \uparrow$ & $\mathtt{Success} \uparrow$ & $\mathtt{BLEU} \uparrow$ \\
\midrule
DAMD \citep{zhang2019task} & 70.00 & 15.00 & 6.90 & 75.00 & 39.50 & 6.20 & 62.50 & 20.50 & 7.60 & 68.00 & 19.50 & 10.50 \\ 
\model{} w/o pre-training & 65.66 & 46.97 & 5.85 & 59.00 & 44.00 & 7.07 & 62.50 & 40.00 & 7.70 & 75.50 & 44.50 & 11.00 \\
\model{} & {\bf86.00} & {65.00} & {12.90} & {80.81} & {64.65} & {9.96} & {74.50} & {43.50} & {8.12} & {81.00} & {55.50} & {12.80} \\
\rowcolor{Gray}
\modelp{L} & {\bf 86.00} & {\bf 68.00} & {\bf 14.60} & {\bf 81.31} & {\bf 74.24} & {\bf 11.90} & {\bf 75.00} & {\bf 51.50} & {\bf 10.09} & {\bf 84.00} & {\bf 62.50} & {\bf 13.17} \\

    \bottomrule
    \end{tabular}
    }
    \caption{End-to-end evaluation on MultiWOZ in the few-shot fine-tuning setting.}
    \label{table:fewshotmultiwoz}
\end{table*}

\begin{table*}[htbp]
    \centering
    \tiny
    
    \setlength{\tabcolsep}{1.0mm}{
    \begin{tabular}{lccccccccccccccc}
    \toprule
    
    \multirow{2}{*}{Model} &
\multicolumn{3}{c}{\texttt{1\%}} &
\multicolumn{3}{c}{\texttt{5\%}} &
\multicolumn{3}{c}{\texttt{10\%}} &
\multicolumn{3}{c}{\texttt{20\%}} \\
\cmidrule(l){2-4} \cmidrule(l){5-7} \cmidrule(l){8-10} \cmidrule(l){11-13} \cmidrule(l){14-16} 

&$\mathtt{Inform} \uparrow$ & $\mathtt{Success} \uparrow$ & $\mathtt{BLEU} \uparrow$ & 
$\mathtt{Inform} \uparrow$ & $\mathtt{Success} \uparrow$ & $\mathtt{BLEU} \uparrow$ & 
$\mathtt{Inform} \uparrow$ & $\mathtt{Success} \uparrow$ & $\mathtt{BLEU} \uparrow$ & 
$\mathtt{Inform} \uparrow$ & $\mathtt{Success} \uparrow$ & $\mathtt{BLEU} \uparrow$ \\
\midrule
DAMD \citep{zhang2019task} & 34.40 & 9.10 & 8.10 & 52.50 & 31.80 & 11.60 & 55.30 & 30.30 & 13.00 & 62.60 & 44.10 & 14.90 \\
\model{} w/o pre-training & 46.10 & 24.40 & 10.39 & 63.40 & 38.70 & 11.19 & 64.90 & 44.50 & 13.57 & 70.10 & 52.20 & 14.72 \\
\rowcolor{Gray}
\model{} & {\bf58.40} & {\bf35.30} & {\bf10.58} & {\bf69.30} & {\bf52.30} & {\bf11.80} & {\bf69.90} & {\bf51.90} & {\bf14.60} & {\bf74.00} & {\bf60.10} & {\bf15.24} 
\\

    \bottomrule
    \end{tabular}
    }
    \caption{End-to-end evaluation on MultiWOZ with varying sizes of task-specific training data for fine-tuning.}
    \label{table:multiwozdiffdata}
\end{table*}

\subsection{Few-Shot Evaluation}
\label{sec:fewshot}
It is desirable for task bots to effectively generalize to new tasks with few task-specific training samples. Thus, the few-shot fine-tuning setting is a more realistic setting for evaluating dialog systems. Unfortunately, the existing task-oriented dialog benchmarks typically contain for each task hundreds to thousands of dialogs.
Therefore, we re-organize CamRest676 and MultiWOZ to simulate the few-shot fine-tuning setting for end-to-end evaluation\footnote{We will release the re-organized datasets.}. We sample from the MultiWOZ dataset the dialog tasks that contain only one domain. \texttt{Attraction}, \texttt{Train}, \texttt{Hotel}, and \texttt{Restaurant} domains are used. We do not use the domains of \texttt{Police}, \texttt{Taxi} and \texttt{Hospital}, as they do not require explicitly tracking dialog states for task completion. For each domain, we randomly sample 50 dialog sessions for training and validation and 200 dialog sessions for testing. 
The only exception is the \texttt{Attraction} domain, which has 100 sessions for testing. 
For CamRest676, we randomly sample 20 sessions. 
Details are shown in Table \ref{table:fewshotdataset}. 

Tables \ref{table:fewshotcamrest} and \ref{table:fewshotmultiwoz} report the end-to-end performance in the few-shot fine-tuning settings on CamRest676 and MultiWOZ, respectively.
On all the domains, \model{}  obtains substantially better performance in all the metrics. Removing task-grounded pre-training significantly hurts the performance of \model{}, although \model{} without task-grounded pre-training still consistently outperforms DAMD in all the domains. \model{} without task-grounded pre-training is conceptually similar to \citet{Ham2020e2e}, but is architecturally simpler and needs fewer annotations. 
The result verifies the importance of task-grounded pre-training on annotated dialog corpora, allowing \model{} to learn how to track dialog and database states to accomplish a task. To study the impact of using larger model size, we build a large version of \model{}, \modelp{L}, which is task-grounded pre-trained on the same data but using GPT-2$_{\text{medium}}$ with 345M parameters as initialization. \modelp{L} consistently outperforms \model{} by a large margin. It indicates that a larger model is a better few-shot learner, exhibiting stronger generalization ability with limited in-domain data. We leave it to future work to significantly scale up \model{}.

We conduct experiments to fine-tune \model{} by varying the percentage of task-specific training samples, ranging from 1\% (80 examples) to 20\% (1600 examples), on the MultiWOZ dataset. 
As shown in Table \ref{table:multiwozdiffdata}, \model{} consistently outperforms DAMD for a wide range of dataset sizes, and the improvement is more substantial when smaller numbers of in-domain examples are used for fine-tuning.

\subsection{Machine Teaching Results}

 \begin{table*}[!h]
    \centering
    \scalebox{0.70}{
    \setlength{\tabcolsep}{1.0mm}{
    \begin{tabular}{lccccccccccccccc}
    \toprule
    
    \multirow{2}{*}{Model} &
\multicolumn{3}{c}{\texttt{Attraction}} &
\multicolumn{3}{c}{\texttt{Train}} &
\multicolumn{3}{c}{\texttt{Hotel}} &
\multicolumn{3}{c}{\texttt{Restaurant}} \\
\cmidrule(l){2-4} \cmidrule(l){5-7} \cmidrule(l){8-10} \cmidrule(l){11-13} \cmidrule(l){14-16} 

&$\mathtt{Inform} \uparrow$ & $\mathtt{Success} \uparrow$ & $\mathtt{BLEU} \uparrow$ & 
$\mathtt{Inform} \uparrow$ & $\mathtt{Success} \uparrow$ & $\mathtt{BLEU} \uparrow$ & 
$\mathtt{Inform} \uparrow$ & $\mathtt{Success} \uparrow$ & $\mathtt{BLEU} \uparrow$ & 
$\mathtt{Inform} \uparrow$ & $\mathtt{Success} \uparrow$ & $\mathtt{BLEU} \uparrow$ \\
\midrule

\model{} & 45.00	&	19.00	&	7.67	&	67.68	&	58.08	&	7.13	&	33.50	&	{\bf22.50}	&	{\bf8.70}	&	50.50	&	10.00	&	8.61 \\
\model{} +\! Extra & 63.00	&	41.00	&	11.08	&	65.15	&	57.58	&	{\bf9.74}	&	41.50	&	19.00	&	7.96	&	44.50	&	27.00	&	9.77 \\
\rowcolor{Gray}
\model{} +\! Teach & {\bf78.00}	&	{\bf45.00}	&	{\bf11.90}	&	{\bf68.18}	&	{\bf63.64}	&	9.45	&	{\bf46.50}	&	{\bf22.50}	&	7.68	&	{\bf53.00}	&	{\bf32.00}	&	{\bf9.81} \\

    \bottomrule
    \end{tabular}}
    }
    \caption{Machine teaching results. \model{} is trained with 10 examples for each domain. {\model{}+Teach} indicates continual training with 5 dialogs recommended by CL with human teacher corrections. {\model{}+Extra} indicates continual training using 5 randomly sampled dialogs with full annotations.  }
    \label{table:humanteaching}
\end{table*}

The machine teaching module of Conversational Learner (CL)~\citep{shukla2020conversation} allows human teachers (dialog authors) to select and visualize dialogs, find potential problems, and provide corrections or additional training samples to improve the bot’s performance. We use CL to evaluate the effectiveness of machine teaching for task bot fine-tuning. In our experiment, 
we first sample 10 dialogs from each domain to fine-tune \model{} as described in Section 3.3. The result is presented in the first row of Table \ref{table:humanteaching}. We then deploy the model to interact with human users via CL. The row of {\model{}+Teach} shows the result of machine teaching, where a human teacher has manually corrected 5 dialogs, which are recommend by CL using a ranking heuristics based on perplexity. The corrections are utilized to continually fine-tune the deployed system.

\begin{figure}[t]
\centering
\includegraphics[width=0.95\columnwidth]{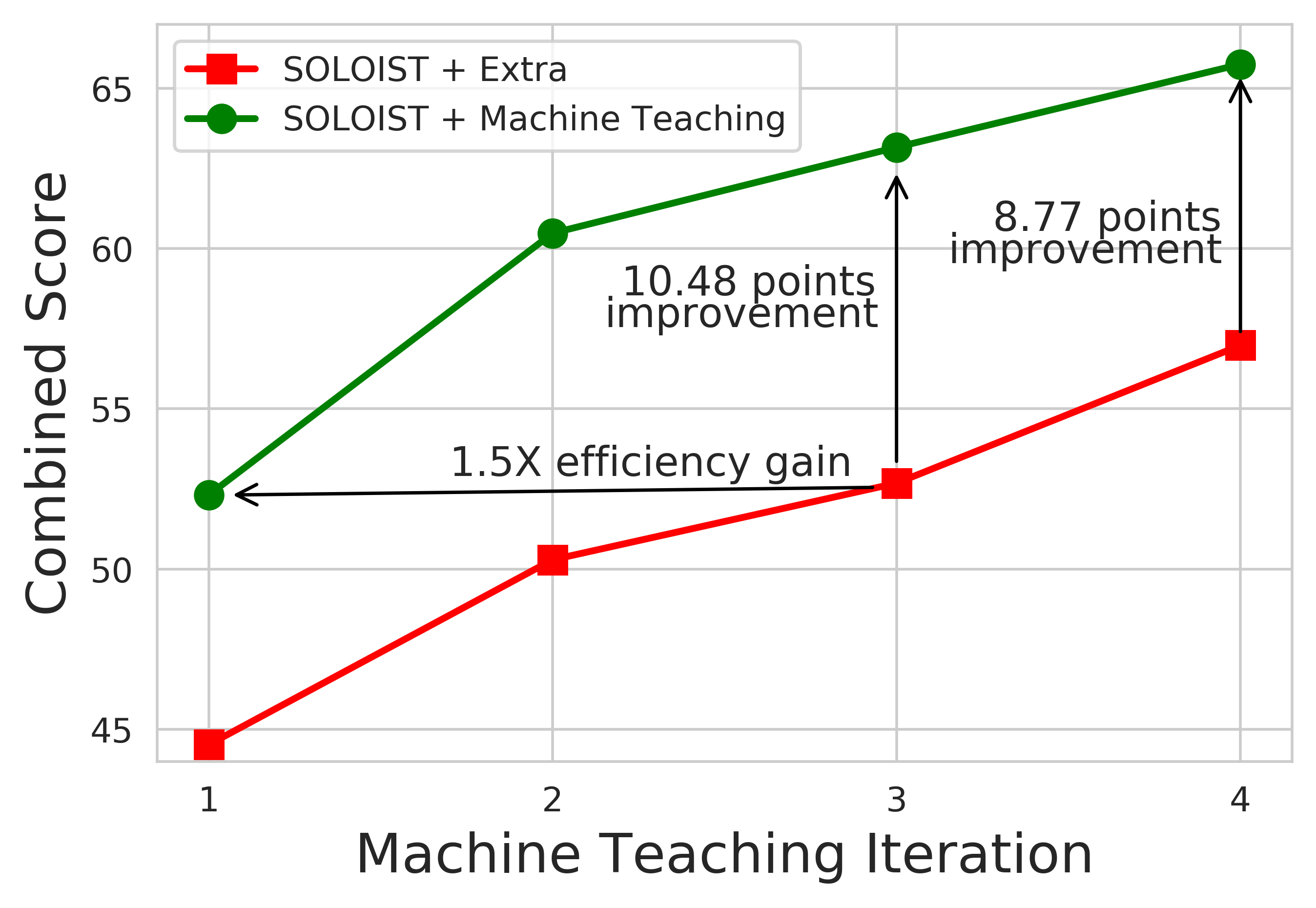}
\caption{Machine teaching performance of different iterations in \texttt{Restaurant} domain. Machine teaching with CL achieves near 1.5X efficiency gain (\ie the 1st iteration used 15 dialogs while the 3rd iteration has 25 dialogs) and boosts performance by 10 points compared with that without teaching.}
\label{fig:mt_curve}
\vspace{-5mm}
\end{figure} 

Table \ref{table:humanteaching} shows that {\model{}+Teach} consistently improves $\mathtt{Combined}$ by a large margin compared with that without human teaching. 
{\model{}+Extra} is used as an ablation baseline, where 5 randomly selected dialogs with full annotations from experts are added as extra examples to fine-tune the model. It shows lower performance than machine teaching. Figure \ref{fig:mt_curve} demonstrates the performance of \model{} in \texttt{Restaurant} by repeating above machine teaching process in multiple iterations. We observe that in the second iteration of machine teaching {\model{}+Teach} improves $\mathtt{Combined}$ by more than 8 points while {\model{}+Extra} achieves 5 points higher. The result demonstrates the effectiveness of our two-step fine-tuning scheme to deploy \model{} for a new task (domain). In terms of machine teaching cost, taking the \texttt{restaurant} domain as an example, we assume that one slot-value pair of belief state correction counts as one edit and a response correction counts as ten edits. The total numbers of edits for {\model{}+Teach} and {\model{}+Extra} are 61 and 396, respectively, suggesting that machine teaching reduces the labeling cost by 6x.

\subsection{Component-Wise Evaluation}
This section evaluates \model{} on two NLU tasks (\ie intent classification and slot filling), the DST task and the response generation task. We show that although \model{} is an end-to-end dialog model, it also performs well on these component tasks.

\begin{table}[h!]
    \centering
    \scriptsize
    
    \setlength{\tabcolsep}{2.5mm}{
    
    \begin{tabular}{lcccc}
    \toprule
    \multirow{2}{*}{Model} & \multicolumn{3}{c}{\texttt{Banking77}} \\
    \cmidrule(l){2-4}
    & 10 & 30 & Full \\
\midrule
BERT-Fixed	& 67.55	& 80.07	& 87.19	\\
BERT-Tuned	& 83.42	& 90.03	& 93.66	\\
USE	& 84.23	& 89.74	& 92.81	\\
ConveRT	& 83.32	& 89.37	& 93.01	\\
USE+ConveRT	& {\bf 85.19} & {\bf 90.57}	& 93.36	\\
\rowcolor{Gray}
\model{} & 78.73 & 89.28 & {\bf 93.80} \\
    \bottomrule
    \end{tabular}
    }
    \caption{Intent classification accuracy scores (5 runs average) on Banking77 with varying number of training examples (10, 30 examples for each intent, and full training examples. The baseline results are cited from \citet{casanueva2020efficient}.}
    \label{table:nlu_banking}
    \vspace{-2mm}
\end{table}

\paragraph{Intent Classification} The task is to classify a user utterance into one of several pre-defined classes (intents). We follow the experiment setting of \citet{casanueva2020efficient}. The last hidden state of \model{} is used as the sequence representation for classification. Several baseline methods are used for comparison. BERT-fixed and BERT-tuned are fine-tuned on BERT, with BERT parameters fixed and updated during fine-tuning, respectively. A linear classifier with a softmax layer is added on top of BERT for classification. Universal Sentence Encoder and ConveRT are sentence encoders tailored for modeling sentence pairs, and are trained for optimizing the conversational response selection task. 
The results in Table \ref{table:nlu_banking} show that \model{} is comparable with SoTA intent classification models. \model{} is the best performer when the full dataset is used for fine-tune but its performance deteriorates more quickly than USE+ConveRT when fewer samples are used for fine-tuning. It is interesting to investigate whether incorporating intent classification tasks in task-grounded pre-training can boost \model{}'s performance. We leave it to future work.    

\begin{table}[h!]
    \centering
    \scriptsize
    
    \setlength{\tabcolsep}{1.0mm}{
    
    \begin{tabular}{l>{\columncolor[gray]{0.95}}cccc}
    \toprule
    
Fraction & \model{} & Span-ConveRT & V-CNN-CRF & Span-BERT \\
\midrule
1 (8198) & {\bf 0.98} & 0.96 & 0.94 & 0.93 \\
1/2 (4099) & {\bf 0.95} & 0.94 & 0.92 & 0.91 \\
1/4 (2049) & {\bf 0.93} & 0.91 & 0.89 & 0.88 \\
1/8 (1024) & {\bf 0.89} & {\bf 0.89 } & 0.85 & 0.85 \\
1/16 (512) & {\bf 0.84} & 0.81 & 0.74 & 0.77 \\
1/32 (256) & {\bf 0.79} & 0.64 & 0.57 & 0.54 \\
1/64 (128) & {\bf 0.74} & 0.58 & 0.37 & 0.42 \\
1/128 (64) & {\bf 0.61} & 0.41 & 0.26 & 0.30 \\
    \bottomrule
    \end{tabular}
    }
    \caption{Average F1 scores across all slots for Restaurant-8K with varying training set fractions. Numbers in brackets represent training set sizes. The baseline results are quoted from \citet{DBLP:conf/acl/CoopeFGVH20}.}
    \label{table:nlu_restaurant}
\end{table}

\paragraph{Slot filling.} We follow the experiment setting of \citet{DBLP:conf/acl/CoopeFGVH20} and formulate slot filling as a turn-based span extraction problem. 
The results in Table \ref{table:nlu_restaurant} show that \model{} performs significantly better than the SoTA method Span-ConveRT, a variant of ConveRT designed explicitly for slot filling. The gap is wider when fewer examples are used for training. For example, when 64 samples are used for training, \model{} outperforms Span-ConveRT by 20 points in F1 score.

\begin{table}
\tiny
    \scriptsize
    {
    \setlength{\tabcolsep}{0.5mm}{

         \begin{tabular}{lcc}
        \toprule
        
\multirow{2}{*}{Model} & \multicolumn{2}{c}{$\mathtt{Joint \ Goal\ Accuracy} \uparrow$} \\
\cmidrule(l){2-3}
& \texttt{MWoz2.0} & \texttt{MWoz2.1} \\
        \midrule 
MDBT\cite{ramadan2018large}
  & 15.57 &  - \\
GLAD\cite{zhong2018global}
  & 35.57 &  - \\
GCE \cite{nouri2018toward}
 & 36.27 &  - \\
FJST \cite{eric2020multiwoz}
 & 40.20 & 38.00 \\
HyST \cite{goel2019hyst}
 & 44.24 &  - \\
SUMBT \cite{lee2019sumbt}
 & 46.65 &  - \\
TOD-BERT \cite{Wu2020ToDBERTPN}
 & - & 48.00 \\
Neural Reading \cite{gao2019dialog}
 & 41.10 & - \\
TRADE \cite{wu2019transferable}
  & 48.62 & 45.60 \\
COMER \cite{ren2019scalable}
  & 48.79 & - \\
NADST \cite{le2020non}
 & 50.52 & 49.04 \\
DSTQA \cite{zhou2019multi}
 & 51.44 & 51.17 \\
SOM-DST \cite{kim2019efficient}
& 51.38 & 52.57 \\
DST-Picklist \cite{zhang2019find}
 & \bf{53.30} &  - \\
MinTL \cite{lin2020mintl}
 & 52.10 & 53.62 \\
SST \cite{chen2020schema}
 & 51.17 & 55.23 \\
Tripy \cite{heck2020trippy}
 & - & 55.29 \\
Simple-TOD \cite{hosseini2020simple}
 & - & 55.72 \\
\rowcolor{Gray}
\model{} & 53.20 & \bf{56.85}  \\
\bottomrule 
\end{tabular}
}}
\caption{Dialog state tracking results on MultiWOZ 2.0 and 2.1. }
    \label{table:multiwoz_dst}
\end{table}

\paragraph{Dialog State Tracking.} 
We compare the dialog state tracking capability of \model{} with several strong baselines on MultiWOZ 2.0 and 2.1.
The results in \ref{table:multiwoz_dst} show that 
\model{} achieves the best performance on MultiWOZ2.1 and similar performance to DST-Picklist \cite{zhang2019find}, which requires pre-defined task ontology to guide state tracking.
In comparison with Simple-TOD~\cite{hosseini2020simple} that is based on GPT-2, \model{} obtains 1.13\% higher joint goal accuracy. We attribute the gain to the task-grounded pre-training that equips \model{} with task completion skills including dialog state tracking.

\paragraph{Context-to-Response.} 

\begin{table*}[!h]
    \centering
    \scriptsize
    \scalebox{1}{
    \setlength{\tabcolsep}{1.0mm}{
    \begin{tabular}{lcccccc}
    \toprule
    
    \multirow{2}{*}{Model} &
\multicolumn{2}{c}{Annotations} &
\multicolumn{4}{c}{Evaluation Metrics} \\
\cmidrule(l){2-3} \cmidrule(l){4-7}
 & Belief State & Policy & $\mathtt{Inform} \uparrow$ & $\mathtt{Success} \uparrow$ & $\mathtt{BLEU} \uparrow$ & $\mathtt{Combined} \uparrow$  \\
\midrule
Baseline \cite{budzianowski2018multiwoz} & \checkmark &  & 71.29 & 60.94 & 18.80 & 84.93 \\
TokenMoE \citep{pei2019modular} & \checkmark & & 75.30 & 59.70 & 16.81 & 84.31 \\  
GPT fine-tuning \citep{budzianowski2019hello} & \checkmark & & 70.96 & 61.36 & 19.05 & 85.21 \\  
Structured Fusion \citep{mehri2019structured} & \checkmark & \checkmark & 82.70 & 72.10 & 16.34 & 93.74 \\
LaRL \citep{zhao2019rethinking} & \checkmark &  & 82.80 & 79.20 & 12.80 & 93.80 \\
MD-Sequicity \citep{zhang2019task} & \checkmark & \checkmark & 86.60 & 71.60 & 16.68 & 95.90 \\ 
HDSA \citep{chen-etal-2019-semantically} & \checkmark & \checkmark & 82.90 & 68.90 & {\bf23.60} & 99.50 \\ 
ARDM \citep{wu2019alternating} &  & & 87.40 & 72.80 & 20.60 & 100.70 \\
DAMD \citep{zhang2019task} & \checkmark & \checkmark & 89.20 & 77.90 & 18.60 & 102.15 \\
\rowcolor{Gray}
\model{} & \checkmark & & {\bf89.60} & {\bf79.30} & 18.03 & \bf{102.49} \\
    \bottomrule
    \end{tabular}
    }}
    \caption{Context-to-response evaluation on MultiWOZ.}
    \label{table:c2rmultiwoz}
\end{table*}

In this task systems need to generate responses given the ground-truth belief state and DB search result~\citep{wen2016network}.
The results on MultiWOZ 2.0 are shown in Table \ref{table:c2rmultiwoz}. 
\model{} achieves the best performance in terms of $\mathtt{Inform}$ and $\mathtt{Success}$ but performs slightly worse in $\mathtt{BLEU}$. 
The $\mathtt{Combined}$ score of \model{} is comparable with the current SoTA method DAMD. 
However, DAMD uses the labels of dialog act on both the user and system sides, which demands significantly higher labeling efforts than \model{} for model training. 
HDSA achieves the best $\mathtt{BLEU}$ score. Compared to HDSA, \model{} is much simpler and able to perform better in terms of $\mathtt{Combined}$. 
\model{} outperforms ARDM in $\mathtt{Combined}$. It is worth mentioning that ARDM cannot perform dialog state tracking and thus requires an extra dialog state tracker to accomplish tasks. These results show that \model{} can learn dialog policies accurately and generate natural language responses in the multi-domain scenario. 

\newcommand\MyHead[2]{%
  \multicolumn{1}{c}{\parbox{#1}{\centering #2}}
}
\begin{table}[!h]
\centering
\scalebox{0.90}{
    \scriptsize
    {
    \setlength{\tabcolsep}{1.0mm}{

         \begin{tabular}{lcccc}
        \toprule

        Model & $\mathtt{Success} \uparrow$ & $\mathtt{Under.} \uparrow$ & $\mathtt{Appr.} \uparrow$ & $\mathtt{Turns} \downarrow$ \\
        \midrule
\rowcolor{Gray}
\model{} & \bf{91.67} & \bf{4.29} & {\bf4.43} & 18.97 \\
DSTC8 Track 1 Winner &  68.32 &   4.15 &  4.29 &   19.51\\
DSTC8 2nd Place & 65.81 & 3.54 & 3.63 & 15.48 \\
DSTC8 3rd Place & 65.09 & 3.54 & 3.84 & {\bf13.88} \\
DSTC8 Baseline & 56.45 & 3.10 & 3.56 & 17.54  \\
\bottomrule \\
\end{tabular}
}}}
\caption{Human evaluation results. The results except \model{} are quoted from~\citet{li2020results}. 
}
    \label{table:human_eval}

\end{table}

\subsection{Human Evaluation Results}
We conduct human evaluation to assess the quality of \model{} interacting with human users. Following the evaluation protocol in DSTC8 track 1 challenge~\cite{kim2019eighth}, we host the best performed \model{} on validation set in MultiWOZ domain in the back-end as bot services and crowdsource the work to Amazon Mechanical Turk. 
For each dialog session, we present Turks a goal with instructions. Then Turks are required to converse with \model{} to achieve the goal and judge the overall dialog experience at the end of a session using four metrics. $(\RN{1})$ $\mathtt{Success}$ evaluates task completion. $(\RN{2})$ $\mathtt{Under.}$ (language understanding score) ranging from 1 (bad) to 5 (good) indicates the extent to which the system understands user inputs. $(\RN{3})$ $\mathtt{Appr.}$ (response appropriateness score) scaling from 1 (bad) to 5 (good) denotes whether the response is appropriate and human-like. $(\RN{4})$ $\mathtt{Turns}$ is the average number of turns in a dialog over all successful dialog sessions.
Turks are further required to write down a justification of giving a specific rating. In total, 120 dialog sessions are gathered for analysis.


Table \ref{table:human_eval} shows the human assessment results on MultiWOZ. The results are consistent with the automatic evaluation. \model{} achieves substantially better performance than other systems over all the metrics. Moreover, \model{} outperforms the DSTC8 Track 1 Winner by a much larger margin in $\mathtt{Success}$ (+20 points) in human evaluation than that in automatic evaluation (+10 points in Figure~\ref{table:e2emultiwoz}).
We attribute this to the fact that Turks use more diverse language to interact with the target bots in interactive human evaluation than that in the pre-collected MultiWOZ dataset and the use of heterogeneous dialog data for task-grounded pre-training makes \model{} a more robust task bot than the others. 
In many test cases against \model{}, Turks comment that they feel like they are talking to a real person.

\begin{figure}[!htbp]
\centering
\includegraphics[width=1\columnwidth]{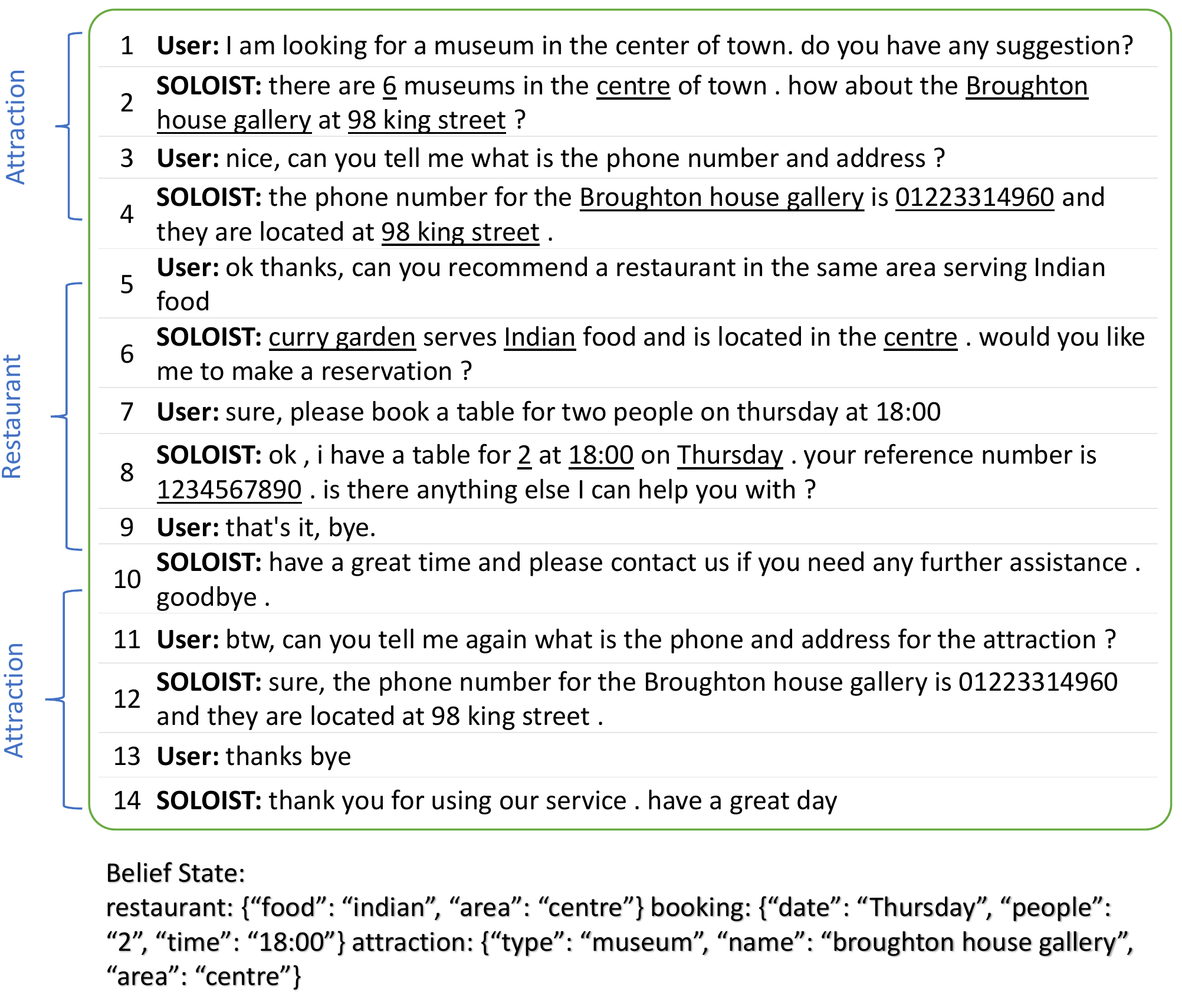}
\caption{An interactive example.}
\label{fig:example}
\end{figure} 

Figure \ref{fig:example} depicts a dialog example where a user interacts with \model{} to complete a multi-domain task.
The user starts the conversation by asking for a recommendation of a museum in the center of town. \model{} identifies the user intent, and provides a recommendation based on the search result from an attraction DB. Then, the user wants to book a table in a restaurant in the same area. 
We can see that through the conversation, \model{} develops belief state, which can be viewed as the system’s understanding of what the user needs and what is available in the DB. Based on the belief state and DB state, \model{} picks the next action, either asking for clarification or providing the user with information being requested. 
This example also demonstrates that \model{} is able to deal with some NLU challenges displayed often in human conversations, such as co-reference resolution. For example, \model{} understands that the ``same area'' at Turn 5 refers to ``centre of town'', and then identifies a proper entity from the restaurant booking DB to make the reservation.

\section{Related Work}

\paragraph{Dialog Systems.}
Dialog systems are typically grouped into two categories, task-oriented systems and social chatbots  ~\citep[\eg][]{chen2017survey,gao2019neural,DBLP:journals/corr/abs-2006-12442,DBLP:journals/coling/ZhouGLS20}.
Recently many variants have been developed to extend the scope of dialog systems, including empathetic dialog systems~\cite{ma2020survey, DBLP:conf/aaai/ZhouHZZL18}, chatbots for sentiment analysis~\cite{li2020bieru}, dialog systems with commonsense knowledge~\cite{young2018augmenting, DBLP:conf/acl/ShusterJRDBW20} or using audio features ~\cite{young2020dialogue}.
In this paper we focus on end-to-end dialog models for task-oriented systems.

\paragraph{Pre-Trained Language Models.}
Recent advances on self-supervised learning have witnessed the blooming of large-scale pre-trained language models~\cite[\eg][]{devlin2019bert,gpt2,dong2019unified}, which achieve SoTA performance on a variety of language understanding and generation tasks. The closest to \model{} are GPT-2~\cite{gpt2} and its variants that ground language generation in the prescribed control codes such as CTRL~\cite{keskar2019ctrl} and Grover~\cite{zellers2019defending}, or latent variables such as Optimus~\cite{li2020optimus}.

Recently, pre-trained language models have been adopted to develop task-oriented and chit-chat dialog systems. 
To name a few examples of chit-chat dialog systems. 
DialoGPT~\cite{zhang2019dialogpt}, TransferTransfo~\cite{wolf2019transfertransfo}
and CGRG~\cite{wu2020Controllable} adapt GPT-2 using human conversational data for response generation. 
Plato~\cite{bao2019plato} pre-trains a discrete latent variable model for response generation. Meena~\cite{adiwardana2020towards} and BST~\cite{roller2020recipes} pre-train large models on conversational data and have demonstrated expressive performance in generating social chit-chat dialogs. 
For task-oriented dialogs, \citet{mehri2019pretraining} explores different  pre-training methods for dialog context representation learning. TOD-BERT~\cite{Wu2020ToDBERTPN} adapts the pre-trained BERT to achieve strong performance on four dialog sub-tasks. 
ConveRT~\cite{DBLP:conf/emnlp/HendersonCMSWV20} pre-trains a model on Reddit data for intent classification and response selection. 
Span-ConveRT~\cite{DBLP:conf/acl/CoopeFGVH20} extends the framework to entity extraction. 
SC-GPT~\cite{peng2020few} uses a pre-trained language model to convert a dialog act to a natural language response. 
All these works use the pre-training and fine-tuning framework. However, they follow the modular architecture of task bots, and the pre-trained models are used for improving individual dialog modules such as NLU and DST. 
\model{} generalizes these methods to the entire dialog pipeline, building an end-to-end dialog system.

\paragraph{End-to-End Trainable Dialog Systems.} 
The end-to-end dialog systems based on neural models have been studied in~\citet{wen2016network,li2017end,lei2018sequicity,haotian2019end}. 
Although these methods have achieved promising results, they are designed for specific domains, rendering difficulties in generalizing to multi-domains such as MultiWOZ. 
Dialog models that can handle multi-domain tasks are studied in 
~\cite{pei2019modular,budzianowski2019hello,mehri2019structured,zhao2019rethinking,wu2019alternating,zhang2019task,peng2017composite}. 
However, these works require large amounts of in-domain labels to achieve good performance.
In contrast, \model{} can effectively adapt to a new task in the few-shot fine-tuning settings.

The most related work to ours is~\citet{Ham2020e2e}, which is the first attempt to fine-tune GPT-2 to build end-to-end dialog models. \citet{hosseini2020simple}  takes a similar approach, and is a concurrent work of \model{}. 
However, \model{} differs from these two methods in two major aspects.
The first is the use of task-grounded pre-training that allows \model{} to learn primary task completion skills, such as tracking dialog states and select system actions. These skills can be easily reused and adapted (e.g., via few-shot fine-tuning) to solve new dialog tasks, leading to a much higher task success rate, as reported in Section 3. 
The second is that the annotation cost required for training \model{} is much lower than that of \citet{Ham2020e2e,hosseini2020simple}.
Training \model{} requires only belief states as labels. But training of \citet{Ham2020e2e,hosseini2020simple} requires labeling each dialog turn with dialog acts.
In addition, while \model{} is end-to-end trainable, the other two models are not and need e.g., heuristic rules to handle different database search conditions. 

\section{Conclusion}
\model{} is a method of building task bots at scale with transfer learning and machine teaching.
Unlike GPT-2, \model{} is pre-trained in a task-grounded manner. So, it can generate responses grounded in user goals and real-world knowledge for task completion. 
Experiments show that \model{} creates new SoTA on two popular task-oriented dialog benchmarks, and that   
\model{} outperforms existing methods by a large margin in the few-shot fine-tuning settings where only a limited amount of task labels are available for fine-tuning.  

We hope that \model{} can inspire dialog researchers and developers to comprehensively explore the new paradigm for building task bots based on task-grounded pre-training and fine-tuning via machine teaching, and improving the recipe we present in this paper, \ie formulating task-oriented dialog as a single auto-regressive language model, pre-training a task-grounded response generation model on heterogeneous dialog corpora, and adapting the pre-trained model to new tasks through fine-tuning using a handful task-specific examples via machine teaching.


\bibliography{tacl}
\bibliographystyle{acl_natbib}

\end{document}